\documentclass[runningheads]{llncs}
\usepackage{amsmath}

\usepackage[T1]{fontenc}
\usepackage{graphicx}
\usepackage[pagebackref,breaklinks,colorlinks]{hyperref}
\usepackage{adjustbox}
\usepackage{multirow}
\usepackage{graphicx}
\usepackage{subcaption}

\begin{document}
\title{Rethinking the Pointer Loss in Table Structure Recognition: Geometry-Aware Pointer Loss for Spatial Locality}
\titlerunning{Geometry-Aware Pointer Loss for Spatial Locality}
%
\author{Hong-Jun Choi\inst{1}\orcidID{0000-0002-3413-511X} \and
Jongho Lee\inst{1}\orcidID{0000-0002-8520-2722} \and
Jaeyoung Kim\inst{1}\orcidID{0000-0003-0880-0398}\thanks{Corresponding author}}

\authorrunning{Choi et al.}

\institute{Teamreboott Inc., South Korea \\
\email{\{hongjun.choi,jongho.lee,jaeyoung.kim\}@reboott.ai}}

\maketitle              

\begin{abstract}
Table Structure Recognition (TSR) using a pointer network achieves impressive results by predicting HTML sequences while aligning tags to detected text (or cell) regions. However, our analysis reveals that when pointer networks fail, 79.6\% of errors occur between spatially adjacent cells (Manhattan distance $\leq$ 2). Despite this, standard cross-entropy loss weights all negative candidates equally. 
In this work, we propose Geometry-Aware Pointer (GAP) Loss, which reweights the cross-entropy objective based on spatial proximity to ground truth. By applying inverse distance weighting, GAP focuses gradient flow where the model struggles most—immediate neighbors receive stronger gradients than distant cells. Our approach requires only a straightforward modification to the loss computation, maintaining the same model architecture with zero additional inference cost.
Extensive experiments on PubTabNet and SynthTabNet demonstrate that GAP consistently reduces adjacent-cell errors, achieving new state-of-the-art performance. Our findings suggest that incorporating geometric inductive biases at the loss level provides a simple yet effective approach to robust TSR. 
Our code is available at \url{https://github.com/teamreboott/GAP}


\end{abstract}
\section{Introduction}
\label{sec:intro}

\begin{figure}[t]
  \centering
  \includegraphics[width=0.55\linewidth]{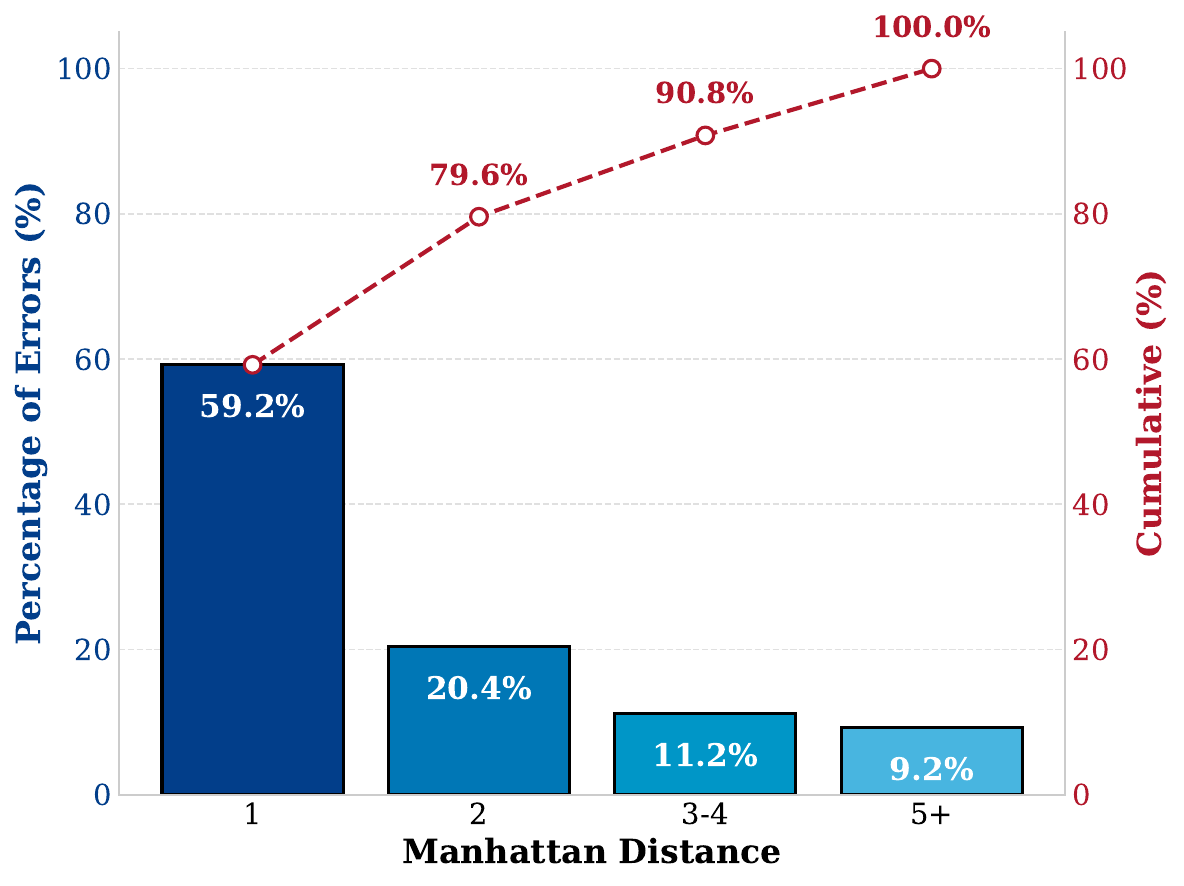}
    \caption{\textbf{Pointer error locality.} Histogram of errors by Manhattan distance \(d\) between the predicted and ground-truth tags (bars), with cumulative percentage (green). \(\,59.2\%\) of errors occur at \(d=1\), \(\,79.6\%\) within \(d\le2\), and \(>\!90\%\) within \(d\le4\).}

  \label{fig:fig1a}
\end{figure}

Tables encode structured information critical for data analysis, financial reporting, and scientific communication. While humans effortlessly parse complex table layouts, automated Table Structure Recognition (TSR) remains challenging due to diverse visual styles, irregular layouts, and ambiguous cell boundaries. The ability to accurately convert table images into machine-readable formats directly impacts downstream applications from information retrieval to automated analytics.

Recent TSR systems have converged on pointer-based architectures~\cite{khang2024tflop,huang2023vast,nassar2022tableformer} that jointly predict structural tags (HTML/OTSL~\cite{lysak2023optimized}) and align them to detected text regions. These models achieve state-of-the-art results by decoupling structure prediction from content localization; a transformer decoder generates the logical structure while a pointer mechanism maps each predicted cell to its corresponding text bounding box. This design enables end-to-end training without complex post-processing rules, making these approaches both effective and practical.

However, the pointer network optimizes a softmax-based pointer loss over the candidate set, where all negative candidates receive uniform treatment during training. This uniform weighting becomes problematic in realistic operating conditions where tables present many closely spaced candidates. When optimizing under this regime, the model struggles to distinguish between spatially adjacent cells, leading to systematic near-miss predictions on neighboring cells instead of the correct one.

To understand this phenomenon, we analyze pointer mistakes as grid offsets in reconstructed tables. For each misprediction, we compute the Manhattan distance $d$ between predicted and ground-truth cells in the table grid. As shown in Figure~\ref{fig:fig1a}, errors exhibit striking locality: 59.2\% occur at $d=1$ (immediate neighbors), 79.6\% within $d \leq 2$, and over 90\% within $d \leq 4$. This concentration reveals a fundamental mismatch: the standard pointer loss assigns equal importance to all negative candidates, yet errors overwhelmingly occur among spatial neighbors.

This observation motivates our central question: \textit{Can we make the pointer loss geometry-aware, emphasizing discrimination between adjacent cells where errors actually occur?} To answer this question, we introduce Geometry-Aware Pointer (GAP) Loss, which reweights the cross-entropy objective based on spatial proximity to ground truth. 
Unlike architectural modifications~\cite{ma2023robust,wang2023tsrformer} that add complexity, or post-processing method~\cite{zhang2022split} that operate outside the learning loop, GAP is a simple change to the loss computation. This modification preserves the original model architecture and adds zero inference cost.
By applying inverse distance weighting, GAP concentrates gradient updates on adjacent cells where the model actually struggles, rather than diluting gradients across all candidates uniformly.
Our contributions are as follows.

\begin{itemize}
  \item \textbf{Systematic error analysis:} We provide the first dedicated study of pointer network failures in TSR, revealing that 79.6\% of errors occur between adjacent cells. This spatial concentration pattern holds across diverse table complexities and datasets, establishing a clear target for optimization improvements.
  
  \item \textbf{Geometry-aware loss design:} We propose GAP, a simple modification to pointer training that incorporates spatial structure through distance-based reweighting. The method is parameter-free while addressing the identified error patterns.
  
  \item \textbf{State-of-the-art results:} Extensive experiments on PubTabNet and SynthTabNet demonstrate that integrating GAP into pointer networks achieves consistent performance gains while reducing adjacent-cell confusions, validating the effectiveness of geometry-aware training.
\end{itemize}

Additionally, we introduce Position Accuracy (PA) as a stringent complementary metric. PA enforces strict tag-to-cell exactness, providing a rigorous assessment of spatial precision that reveals alignment failures masked by more lenient structural similarity metrics.

\section{Related work}

TSR approaches can be categorized by how they handle the spatial alignment between predicted structure and detected content. We organize prior work based on their alignment strategies and position our geometry-aware loss within this landscape.

\paragraph{Structure-to-Content Alignment Methods.}

Early TSR systems employ detection-based~\cite{qasim2019rethinking,riba2019table,chi2019complicated} or graph-based~\cite{zhang2022split,wang2023tsrformer,xue2021tgrnet} pipelines that explicitly model spatial relationships. However, they suffer from error propagation across stages and require domain-specific components. Recent end-to-end approaches address these limitations: TableFormer~\cite{nassar2022tableformer} and TableMaster~\cite{ye2021pingantablemaster} use transformer decoders with bounding box regression, VAST~\cite{huang2023vast} introduces visual alignment loss, DRCC~\cite{shen2023divide} proposes cascaded decoding for large tables, and SPRINT~\cite{kudale2024sprint} reformulates TSR as language-agnostic cell arrangement and SLANet~\cite{dimitri2025slanet} demonstrates lightweight alternatives using transformer-free architectures. While these methods advance structure prediction, they share a common limitation—the pointer training objective treats all candidates uniformly, which our work directly addresses.

\paragraph{Pointer Networks for TSR.}

Pointer networks~\cite{vinyals2015pointer,khang2024tflop} have emerged as the dominant mechanism for aligning predicted structure tags with detected content regions, computing attention scores between decoder states and candidate bounding boxes. TFLOP~\cite{khang2024tflop} achieves state-of-the-art performance by combining BART~\cite{lewis-etal-2020-bart}-style decoding with pointer attention.

Despite their success, existing methods optimize pointer alignment using standard cross-entropy loss that treats all negative candidates uniformly. This spatial blindness becomes problematic in dense tables where numerous candidates cluster in close proximity—the loss penalizes selection of a distant cell equally to an adjacent cell. Our work addresses this limitation through GAP loss, which reweights training based on Manhattan distance. Unlike architectural modifications~\cite{ma2023robust,wang2023tsrformer}, GAP operates purely at the loss level with zero inference cost, directly targeting the spatial confusion patterns where pointer networks struggle most.

\paragraph{Evaluation Metrics for TSR}

Evaluation metrics for TSR, including TEDS~\cite{zhong2020image}, adjacency-based measures~\cite{gobel2012methodology,gao2019icdar}, and GriTS~\cite{smock2022grits}, primarily evaluate structural coherence rather than exact correspondence between cell content and its spatial position. TEDS, the de facto standard metric benchmark metric, operates on HTML tree similarity where systematic pointer shifts incur only partial penalties as content substitutions. This allows models to achieve high TEDS scores while harboring systematic alignment errors—a critical gap for pointer-based systems where adjacent-cell confusions dominate. 
To address this limitation, we propose Position Accuracy (Section 4.3), which enforces strict cell-level correctness rather than partial structural similarity.

\section{Analysis of Pointer Mechanisms in TSR}
We conduct our analysis on TFLOP, the current state-of-the-art TSR model that exemplifies modern pointer-based approaches. This model uses a transformer decoder for OTSL sequence generation coupled with a pointer mechanism for tag-to-cell (bounding box) alignment. The following subsections first introduce the key components of TFLOP, then present our empirical analysis revealing limitations in current pointer training objectives.

\subsection{Background}

\begin{figure}[t]
  \centering
  \includegraphics[width=0.6\linewidth]{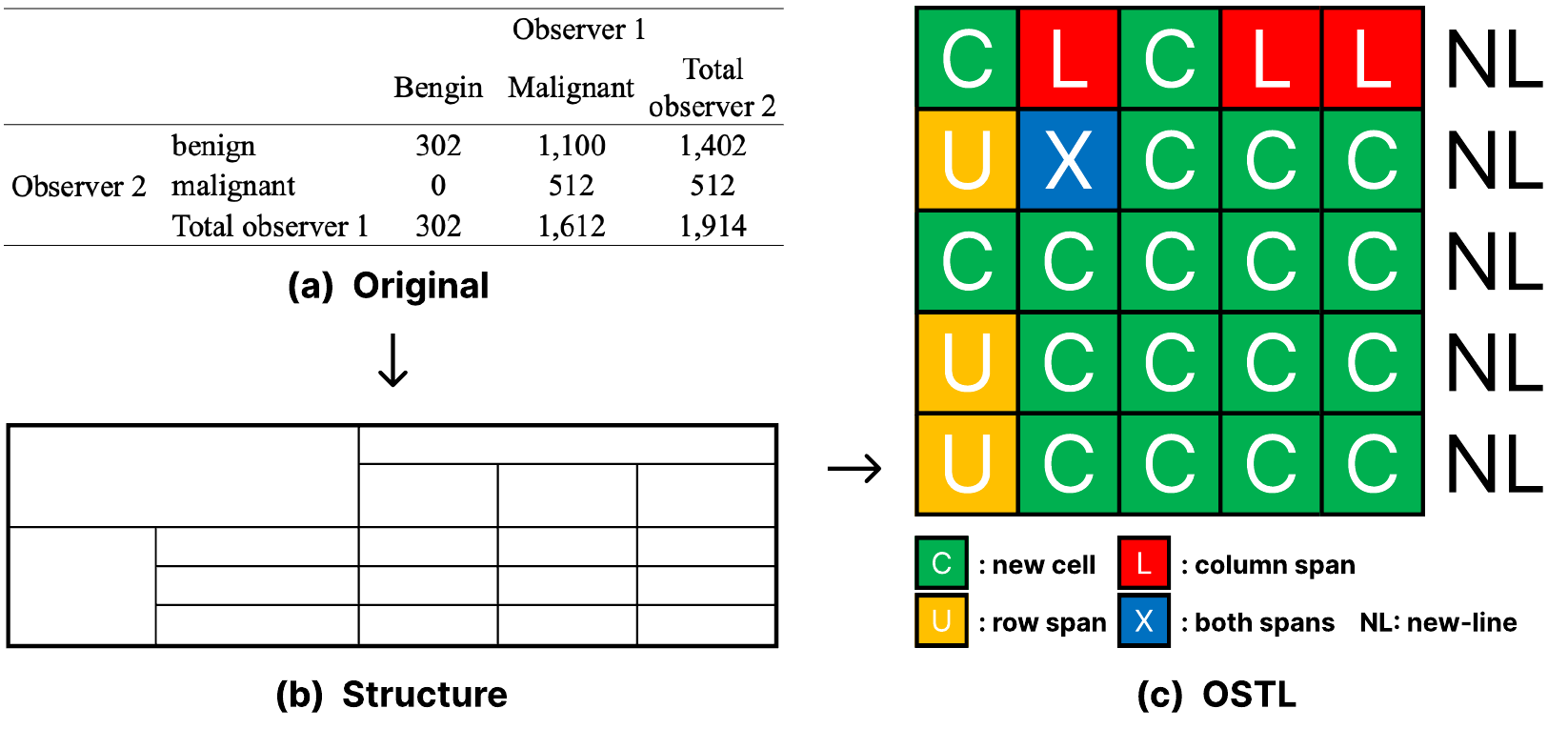}
    \caption{\textbf{OTSL representation of table structure.} From left to right: (a) Original table with detected text regions, (b) Reconstructed grid structure showing cell boundaries and spans, (c) OTSL token sequence in grid format where green \texttt{C} tokens represent cells requiring pointer mapping to text regions, colored tokens (\texttt{L}, \texttt{U}, \texttt{X}) indicate span continuations, and \texttt{NL} marks row endings. The pointer mechanism maps each \texttt{C} token to its corresponding bounding box.}
  \label{fig:fig2}
\end{figure}

\paragraph{Architecture Overview.}
TFLOP processes table images through a Swin Transformer encoder to extract visual features $\{\mathbf{z}_i\}_{i=1}^P$. Subsequently, a layout encoder consisting of multi-layer perceptrons (MLPs) embeds text region bounding boxes by aggregating coordinate embeddings with visual features extracted via ROIAlign~\cite{he2017mask} from $\mathbf{z}$, producing layout embeddings $\{\mathbf{l}_j\}_{j=1}^M$ where $M$ denotes the number of detected text regions.

\paragraph{Structure Generation with OTSL.}
The BART decoder generates OTSL-tag sequences conditioned on visual features and layout embeddings. As illustrated in Figure~\ref{fig:fig2}, OTSL represents table structure using five tokens: \texttt{C} (new cell), \texttt{L} (column span), \texttt{U} (row span), \texttt{X} (both spans), and \texttt{NL} (newline). This compact representation maps directly to a 2D grid where each \texttt{C} token requires pointer alignment to a specific text region. The decoder is supervised with cross-entropy loss for structure generation:
\begin{equation}
\mathcal{L}_{cls} = -\frac{1}{T}\sum_{k=1}^{T} \log p(y_k^* | y_{<k}, \mathbf{z}, \mathbf{l}),
\end{equation}
where $T$ is the sequence length, $y_k^*$ is the ground-truth tag at position $k$, and $y_{<k}$ represents all previously generated tags.

\paragraph{Layout Pointer Mechanism.}
The BART decoder's last hidden states $\{\mathbf{h}_i\}_{i=1}^N$ are split into bounding-box features $\{\mathbf{h}_j^{bbox}\}_{j=1}^M$ and tag features $\{\mathbf{h}_k^{tag}\}_{k=1}^T$, where $N = M + T$. These are projected through separate linear layers: $\bar{\mathbf{h}}_j^{bbox} = \text{proj}_b(\mathbf{h}_j^{bbox})$ and $\bar{\mathbf{h}}_k^{tag} = \text{proj}_t(\mathbf{h}_k^{tag})$. 

The pointer mechanism specifically targets \texttt{C} tokens, as these are the only OTSL tokens that correspond to actual table cells containing text. While structural tokens (\texttt{L}, \texttt{U}, \texttt{X}, \texttt{NL}) define the table layout and cell spans, only \texttt{C} tokens require alignment with detected text regions. For each position $k$ where $y_k = \texttt{C}$, the pointer computes attention scores over all text regions:
\begin{equation}
p(j | k) = \frac{\exp(\bar{\mathbf{h}}_k^{tag} \cdot \bar{\mathbf{h}}_j^{bbox} / \tau)}{\sum_{j'=1}^{M} \exp(\bar{\mathbf{h}}_k^{tag} \cdot \bar{\mathbf{h}}_{j'}^{bbox} / \tau)}
\end{equation}
The pointer loss minimizes the negative log-likelihood of correct alignments:
\begin{equation}
\mathcal{L}_{ptr} = -\frac{1}{|C|}\sum_{k \in C} \log p(j_k^* | k),
\end{equation}
where $C = \{k : y_k = \texttt{C}\}$ denotes the set of positions where \texttt{C} tokens appear, $j_k^*$ is the ground-truth bounding box index for position $k$, and $\tau$ is the temperature parameter.

\paragraph{Training Objective.}
TFLOP's training objective combines multiple loss components:
\begin{equation}
\mathcal{L}_{total} = \mathcal{L}_{cls} + \lambda_1 \mathcal{L}_{ptr} + \lambda_2 \mathcal{L}_{ptr}^{empty} + \lambda_3 \mathcal{L}_{contr},
\label{eq:tflop_total}
\end{equation}
where $\mathcal{L}_{cls}$ supervises structure generation and $\mathcal{L}_{ptr}$ optimizes pointer alignment. The auxiliary losses $\mathcal{L}_{ptr}^{empty}$ and $\mathcal{L}_{contr}$ handle empty cell detection and span-aware contrastive learning respectively (see \cite{khang2024tflop} for implementation details).

\subsection{Spatial Error Analysis}
The OTSL representation naturally maps to a 2D grid, enabling us to quantify spatial relationships between cells. For any two cells $i$ and $j$ at grid positions $(r_i, c_i)$ and $(r_j, c_j)$, we define their Manhattan distance as:
\begin{equation}
d(i,j) = |r_i - r_j| + |c_i - c_j|,
\label{eq:l1_dist}
\end{equation}
where $d(i,j)=1$ represents immediate neighbors (up, down, left, right), $d(i,j)=2$ includes diagonal cells, and so forth. Throughout this paper, we use the shorthand notation $d=k$ to refer to $d(i,j)=k$.

\paragraph{Error Locality.}
Using this distance metric, we analyze pointer prediction errors across the validation set of PubTabNet. As shown in Figure~\ref{fig:fig1a}, errors exhibit striking locality: 59.2\% occur at $d=1$ (horizontally or vertically adjacent cells), 79.6\% within $d \leq 2$, and over 90\% within $d \leq 4$. This concentration persists across different table complexities and sizes, indicating a fundamental challenge in distinguishing spatially proximate cells.

\begin{figure}[t]
\centering
\begin{subfigure}[b]{0.3\linewidth}
    \centering
    \includegraphics[width=\linewidth]{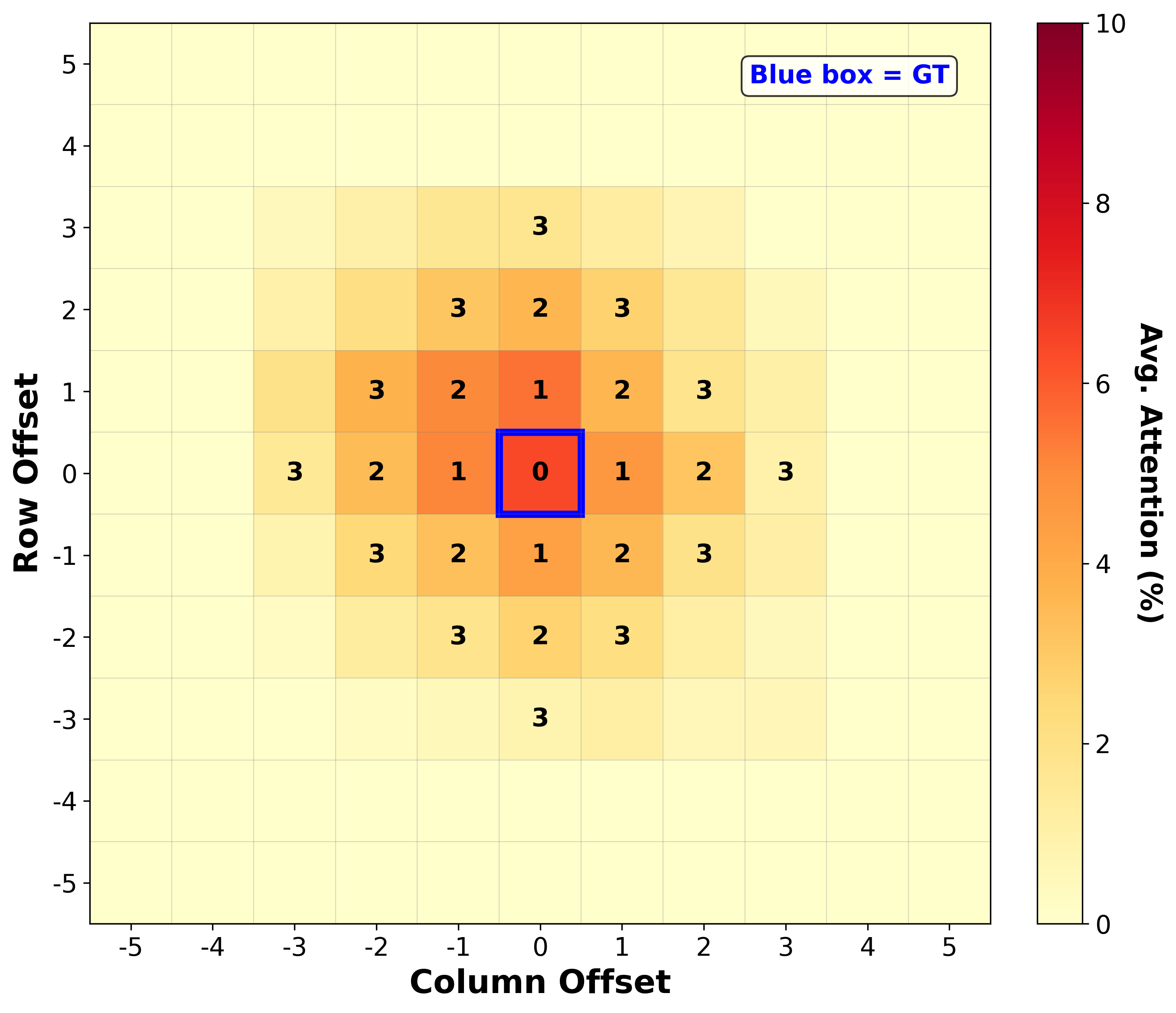}
    \caption{Attention distribution for incorrect predictions}
    \label{fig:attention_dist}
\end{subfigure}
\begin{subfigure}[b]{0.4\linewidth}
    \centering
    \includegraphics[width=\linewidth]{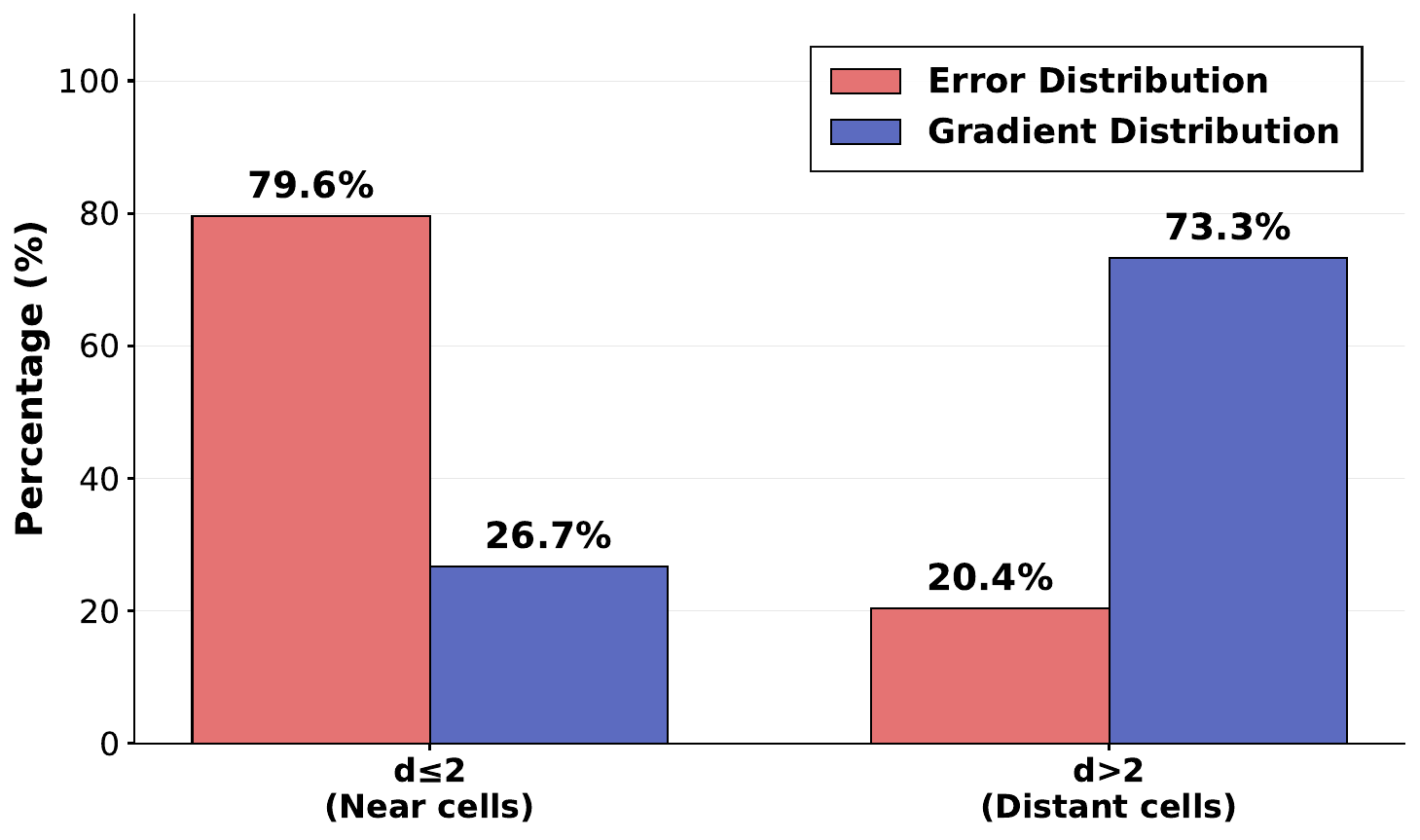}
    \caption{Error-gradient distribution mismatch}
    \label{fig:error_grad_discrepancy}
\end{subfigure}

\caption{\textbf{Analysis of pointer network confusion patterns.} 
(a) Attention heatmap showing that errors concentrate around the correct region: ground truth receives 6.39\% attention while nearby cells ($d=1,2$) collectively capture 47.6\%.
(b) Mismatch between error distribution (79.6\% at $d \leq 2$) and gradient distribution (26.7\% at $d \leq 2$) reveals the spatial blindness of standard cross-entropy loss.}
\label{fig:pointer_analysis}
\end{figure}

\paragraph{Attention Distribution Analysis.}
To understand why these local errors occur, we analyze the attention distribution when the pointer makes mistakes. Specifically, for each incorrect prediction where $j_{pred} \neq j^*$, we extract the attention weights by computing the cosine similarity between the L2-normalized tag features $\bar{\mathbf{h}}_k^{tag}$ and bounding box features $\bar{\mathbf{h}}_j^{bbox}$ obtained from the decoder's linear projections. These similarities are then passed through a softmax function to produce attention probabilities. We group these attention weights by their Manhattan distance $d(j^*, j) = |r_j - r_{j^*}| + |c_j - c_{j^*}|$ from the ground truth position $j^*$, where $(r_j, c_j)$ represents the row and column indices in the table grid.
 
As shown in Figure~\ref{fig:attention_dist}, when the pointer predicts incorrectly, the attention distribution reveals a striking pattern: while the ground truth position receives only 6.39\% attention, cells at Manhattan distance $d=1$ and $d=2$ collectively capture 47.6\% of the total attention mass (19.6\% and 28\% respectively). This concentration of attention around the correct position, rather than precisely on it, suggests that the model successfully identifies the coarse spatial region but struggles with fine-grained localization among neighboring cells.

\paragraph{Learning Inefficiency in Pointer Loss.}
To analyze the inefficiency in standard cross-entropy optimization, we examine gradient distribution for predictions with TEDS $<$ 0.6 from PubTabNet. This threshold excludes trivial cases where the model performs well, focusing on challenging scenarios where the pointer mechanism struggles.

We analyze how gradient is distributed among negative bounding box candidates for each tag-to-bounding box prediction. Specifically, we compute the ratio of gradient magnitudes received by nearby candidates (bounding boxes corresponding to cells within Manhattan distance $d \leq 2$ from the ground-truth cell) to the total gradient across all negative candidates:
\begin{equation}
    \text{Gradient Ratio} = \frac{\sum_{j \in \mathcal{M}, d(c_j, c_{j_k^*}) \leq 2, j \neq j_k^*} |\nabla_j|}{\sum_{j' \in \mathcal{M}, j' \neq j_k^*} |\nabla_{j'}|} \times 100\%,
\end{equation}
where $\mathcal{M}$ denotes the set of all detected bounding boxes, $j_k^*$ is the ground-truth bounding box for C-tag $k$, and $c_j$ represents the cell position in the table grid corresponding to bounding box $j$.

In Figure~\ref{fig:error_grad_discrepancy}, while 79.6\% of actual pointer errors occur at nearby cells ($d \leq 2$), these positions receive only 26.7\% of the total gradient signal. The remaining 73.3\% is dispersed across distant cells that account for merely 20.4\% of errors.

This mismatch stems from the limitation of cross-entropy loss, which treats all negative candidates equally without spatial awareness. For example, among 100+ C-tags in a table, only 4-8 neighboring cells represent genuine confusion risks, yet they receive identical gradient magnitudes as distant cells. Consequently, the model disperses its learning capacity uniformly across all positions rather than developing fine-grained discrimination between spatially proximate cells, where such capability is most critical for accurate TSR.

\begin{figure}[t]
\centering
\begin{subfigure}[t]{0.25\linewidth}
    \centering
    \includegraphics[width=\linewidth]{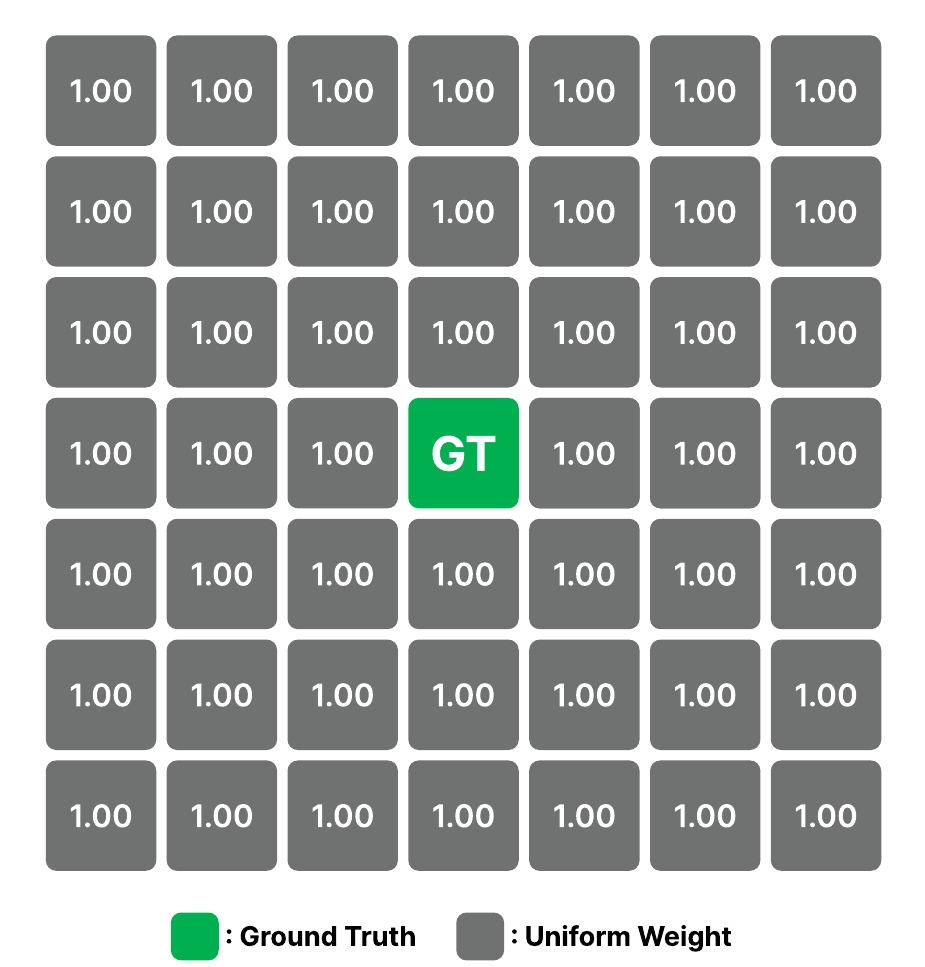}
    \caption{Standard Pointer loss}
    \label{fig:standard_ce}
\end{subfigure}
\begin{subfigure}[t]{0.25\linewidth}
    \centering
    \includegraphics[width=\linewidth]{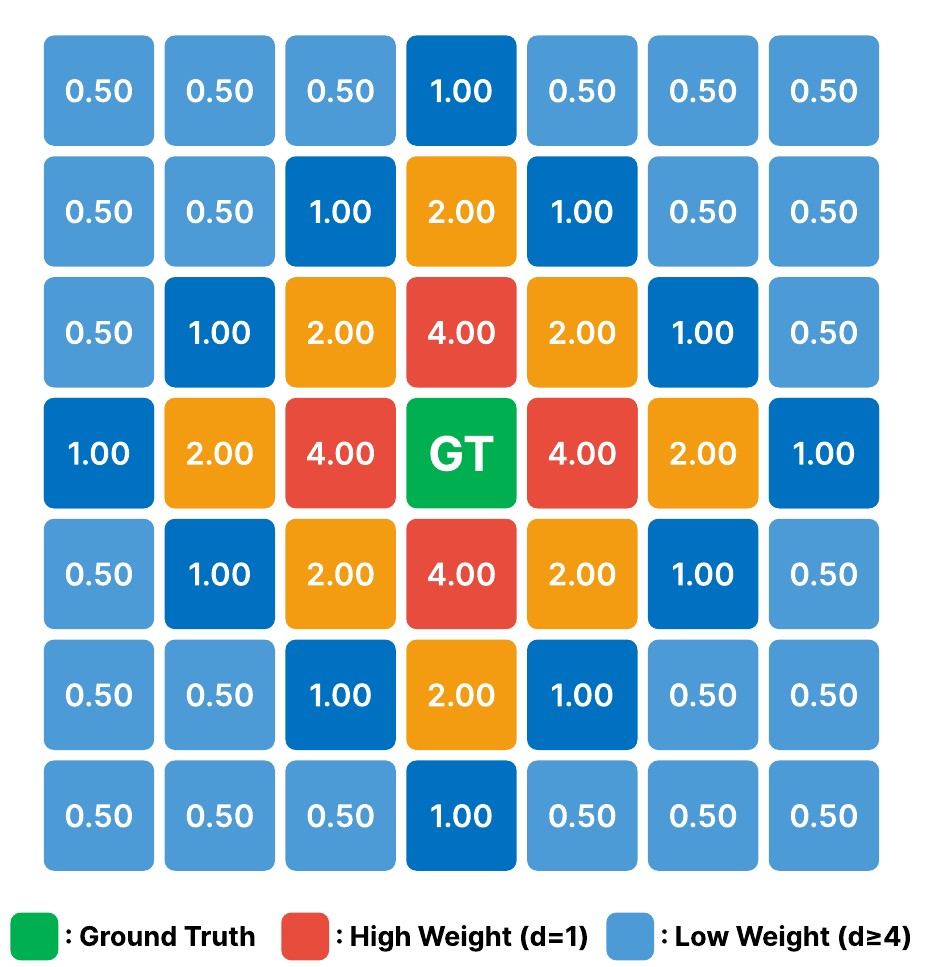}
    \caption{GAP loss}
    \label{fig:gce_weights}
\end{subfigure}
\caption{Weight distribution comparison in a 7×7 table grid. (a) Standard pointer loss assigns uniform weights (w=1.0) to all negative candidates, ignoring spatial structure. (b) Our GAP applies distance-based weighting with geometric decay. The center cell (GT) represents the ground truth position.}
\label{fig:gpl_comparision}
\end{figure}

\section{Methodology}

\subsection{Spatial Weight Design}
Our analysis motivates the GAP loss, a training objective designed to integrate spatial structure by reweighting each negative sample's contribution based on its proximity to the ground-truth cell.

Building upon the Manhattan distance metric, we assign spatial weights that decay exponentially with distance:
\begin{equation}
w(d) = \max\left(\frac{\alpha}{2^{d-1}}, 0.5\right), \quad d \geq 1,
\label{eq:weight}
\end{equation}
where $d = d(c_{j_k^*}, c_j)$ is the Manhattan distance between cells in the table grid, and $\alpha$ controls the base weight scale. The weight applies only to negative candidates as the ground truth does not require reweighting.

Through empirical evaluation with $\alpha \in \{4, 8, 16\}$, we found $\alpha = 8$ achieves the best performance, yielding weights of 4, 2, 1 for distances 1, 2, 3 respectively, then maintaining 0.5 for distant cells. We use $\alpha = 8$ for all experiments. Figure~\ref{fig:gpl_comparision} visualizes this weighting distribution.

\begin{figure}[t]
\centering
\includegraphics[width=0.6\linewidth]{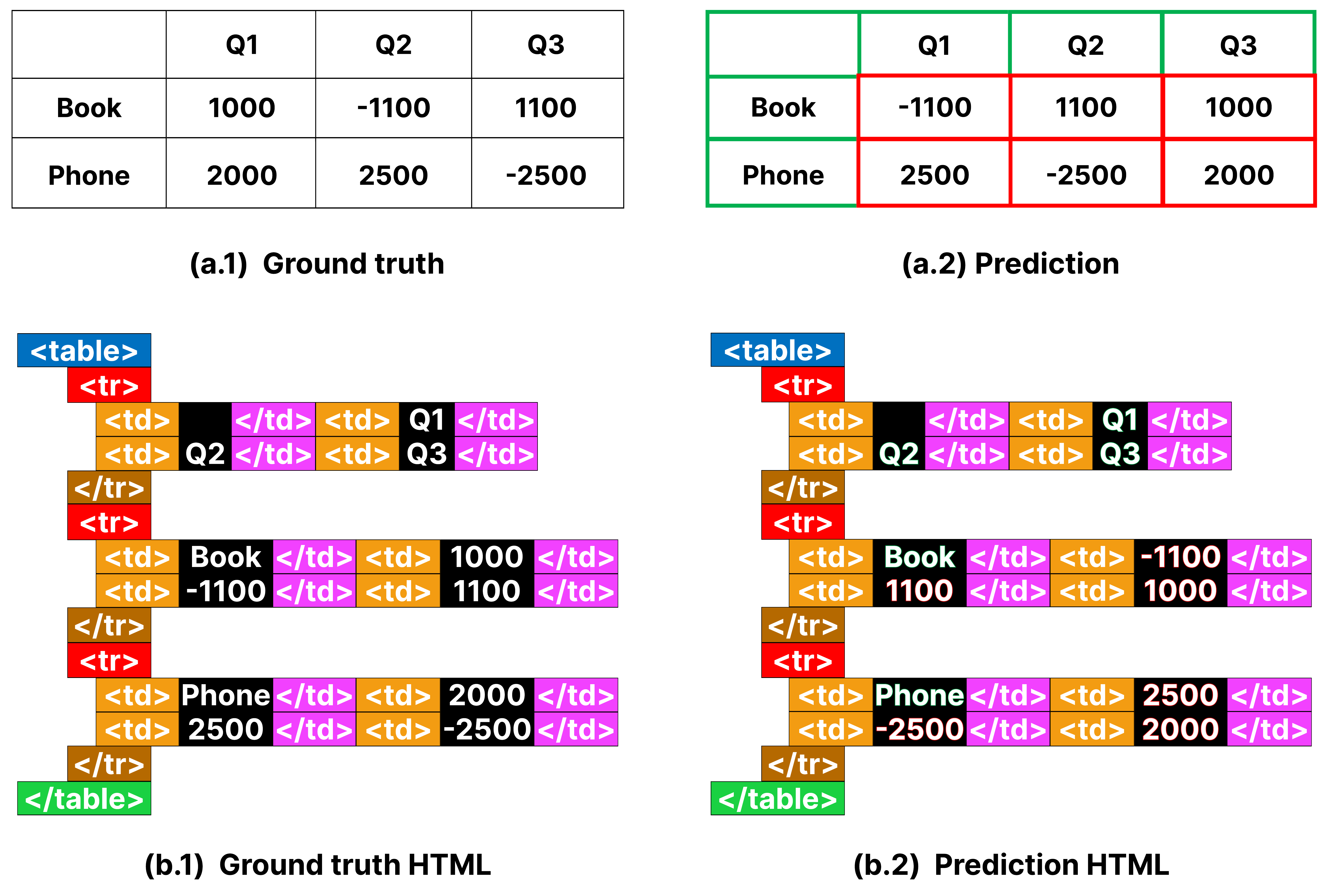}

\caption{\textbf{TEDS vs Position Accuracy comparison.} Ground truth table (left) and model prediction (right) where all data columns are systematically shifted. TEDS-structure achieves 100\% and TEDS evaluates this as 90.0\% due to preserved structural relationships and partial content scores, while Position Accuracy reveals the true severity with only 50.0\% of cells correctly positioned. This discrepancy highlights TEDS's limitation in detecting systematic pointer misalignment.}
\label{fig:teds_vs_pa}
\end{figure}

\subsection{Geometry-Aware Pointer Loss}
The GAP loss incorporates spatial weights into the pointer loss computation. For each C-tag position $k \in C$, the modified loss is:
\begin{equation}
\mathcal{L}_{GAP} = -\frac{1}{|C|} \sum_{k \in C} \log \frac{\exp(s_{j_k^*})}{\exp(s_{j_k^*}) + \sum_{j \neq j_k^*} w_j \exp(s_j)},
\end{equation}
where $s_j = \bar{\mathbf{h}}_k^{tag} \cdot \bar{\mathbf{h}}_j^{bbox}/\tau$ denotes the attention score, $j_k^*$ is the ground-truth bounding box for C-tag $k$, and $w_j = w(d(c_{j_k^*}, c_j))$ is the spatial weight based on Manhattan distance between cells in the table grid.

By intensifying gradients for nearby cells while suppressing distant ones, GAP enables the model to focus its learning capacity on fine-grained discrimination where confusions actually occur.

GAP requires no architectural modifications and operates purely at the loss level. During inference, the model uses the same forward pass and pointer mechanism as the original TFLOP. In the training objective, $\mathcal{L}_{GAP}$ simply replaces the original $\mathcal{L}_{ptr}$ from Equation~\ref{eq:tflop_total}, with all other components remaining unchanged.

\subsection{Evaluation Metric for Pointer Networks}
\subsubsection{Limitations of Existing TSR Metrics}

 Current TSR metrics provide valuable insights but fail to capture critical pointer network failures. As illustrated in Figure 5, systematic cell shifts yield misleadingly high TEDS scores (90.0\%) despite only 50.0\% of cells being correctly positioned. This discrepancy arises because TEDS operates on character-level edit distance, where shifted content incurs minimal penalties as partial substitutions rather than complete errors. From a practical standpoint, a financial table where profits become losses due to column shifts represents complete failure, not a 10\% degradation. This diagnostic blind spot necessitates position-aware evaluation with binary cell-level accuracy—either correct position and content, or complete failure.

\subsubsection{Position Accuracy}
To address these limitations, we propose Position Accuracy (PA), which implements the binary evaluation paradigm by directly measuring pointer accuracy:
\begin{equation}
\label{eq:position_acc}
\text{PA} = \frac{|\{k \in C : j_k^{pred} = j_k^*\}|}{|C|},
\end{equation}
where $C = \{k : y_k = \texttt{C}\}$ denotes the set of C-tag positions in the OTSL sequence, $j_k^{pred}$ is the predicted bounding box index for C-tag $k$, and $j_k^*$ is the ground-truth bounding box index. 

Unlike TEDS's partial scoring, PA treats each cell as either correctly positioned or completely wrong, reflecting the practical reality that mispositioned data is unusable regardless of content similarity. For spanning cells, we evaluate pointer assignment to the anchor position (the C-tag), consistent with training protocols.

\section{Experiments}
\subsection{Datasets and Experimental Configuration}
We conduct experiments on two widely adopted table recognition datasets: PubTabNet~\cite{zhong2020image} and SynthTabNet \cite{nassar2022tableformer}.

\textbf{PubTabNet} represents a comprehensive TSR benchmark featuring HTML-annotated tables sourced from scientific publications. The dataset includes 500,777 training samples and 9,115 validation images, with an official test split containing 9,064 table images. A notable characteristic of the test partition is the absence of ground truth cell-level bounding boxes, necessitating the use of external OCR systems for text region extraction during evaluation.\footnote{Following TFLOP, we utilize the pre-extracted OCR bounding boxes available at \url{https://github.com/UpstageAI/TFLOP} for consistent evaluation on the PubTabNet test set.}

\textbf{SynthTabNet} offers a synthetic approach to table recognition benchmarking, developed to mitigate distributional limitations present in naturally occurring table datasets. This dataset provides 600,000 artificially generated table images distributed across four equal partitions of 150,000 samples each. 

\textbf{Implementation Details}
We follow the experimental configuration of TFLOP~\cite{khang2024tflop} to ensure fair comparison. Input images are rescaled to 768$\times$768 resolution with batch size of 64. The loss function hyperparameters remain unchanged with $\lambda_1 = \lambda_2 = 1.0$ and $\lambda_3 = 0.5$. Lastly, training is performed on 2$\times$H200 GPUs.

\subsection{Evaluation Metrics}
To evaluate GAP loss, we utilize three evaluation metrics.

\textbf{TEDS}: Tree-Edit-Distance-based Similarity measuring structural similarity between predicted and ground truth HTML with content:
\begin{equation}
\text{TEDS}(T_{pred}, T_{gt}) = 1 - \frac{\text{EditDist}(T_{pred}, T_{gt})}{\max(|T_{pred}|, |T_{gt}|)},
\end{equation}
where $T_{pred}$ and $T_{gt}$ denote the predicted and ground truth HTML trees respectively, $\text{EditDist}(\cdot, \cdot)$ computes the tree edit distance, and $|\cdot|$ represents the number of nodes in the tree.

\textbf{TEDS-Struct (TEDS-S)}: TEDS computed without table text content, focusing solely on structural accuracy. This metric uses the same formula as TEDS but operates on structure-only trees where text content is removed.

\textbf{Position Accuracy (PA)}: Metric of pointer precision by evaluating C-tag to bounding box alignment as defined in Equation~\ref{eq:position_acc}.

\subsection{Main Results}

\begin{table}[t]
\centering
\caption{Performance comparison on PubTabNet test set and SynthTabNet validation datasets.}
\label{tab:main_results}
\footnotesize
\setlength{\tabcolsep}{5pt}
\begin{tabular}{l|c|c||c|c}
\hline
\multirow{2}{*}{\textbf{Methods}} & \multicolumn{2}{c||}{\textbf{PubTabNet}} & \multicolumn{2}{c}{\textbf{SynthTabNet}} \\
& \textbf{TEDS-S} & \textbf{TEDS} & \textbf{TEDS-S} & \textbf{TEDS} \\
\hline
TableMaster [2021] & - & 96.32 & - & - \\
TableFormer [2022] & 97.5 & 93.60 & 96.70 & - \\
VAST [2023] & 97.23 & 96.31 & - & - \\
DRCC [2023] & - & - & 98.70 & - \\
SLANET [2025] & 97.48 & 95.89 & - & - \\
SPRINT [2025] & 97.68 & - & 99.36 & - \\
\hline
TFLOP & \underline{98.27} & \underline{96.43} & \underline{99.53} & \underline{99.25} \\
\hline
TFLOP + GAP (Ours) & \textbf{98.28} & \textbf{96.49} & \textbf{99.69} & \textbf{99.53} \\
\hline
\end{tabular}
\end{table}

\begin{table}[t]
\centering
\caption{Position accuracy (PA) results in PubTabNet and SynthTabNet.}
\label{tab:tca_results}
\footnotesize
\begin{tabular}{l|c|c}
\hline
\textbf{Methods} & \textbf{PubTabNet} & \textbf{SynthTabNet} \\
\hline
TFLOP & 91.80 & 98.75 \\
+ GAP Loss &  \textbf{92.35} & \textbf{99.31} \\
\hline
\end{tabular}
\label{tab:position_acc}
\end{table}

\begin{table}[t]
\centering
\caption{TEDS-PA disagreement analysis on PubTabNet test set. Percentage of high-TEDS samples ($\geq$90\%) with substantial pointer errors.}
\label{tab:disagreement}
\small
\begin{tabular}{l|cc|c}
\hline
\textbf{Misalignment} & \textbf{TFLOP} & \textbf{+GAP} & \textbf{Relative} $\downarrow$ \\
\hline
Moderate (PA $<$ 80\%) & 6.60\% & 5.75\% & 12.9\% \\
High (PA $<$ 70\%) & 3.84\% & 3.33\% & 13.3\% \\
\hline
\end{tabular}
\end{table}

Table~\ref{tab:main_results} demonstrates the effectiveness of our approach across both benchmark datasets. TFLOP achieves strong baseline performance with TEDS scores of 96.43 on PubTabNet and 99.53 on SynthTabNet, confirming the robust pointer network capabilities of the original framework. Replacing TFLOP's standard pointer loss with our GAP loss yields consistent improvements in TEDS scores, achieving 96.49 on PubTabNet and 99.53 on SynthTabNet. 

The PA results presented in Table~\ref{tab:position_acc} reveal more substantial improvements. On PubTabNet, GAP loss enhances PA from 91.80 to 92.35, representing a 0.55 percentage point gain. Similarly, SynthTabNet shows improvement from 98.75 to 99.31, an increase of 0.56 percentage points. These PA enhancements substantially exceed the corresponding TEDS improvements, indicating that our geometry-aware approach specifically addresses pointer precision challenges.

To validate the necessity of position-aware evaluation at corpus level, we analyze samples where TEDS indicates strong performance but PA reveals significant pointer errors. As shown in Table~\ref{tab:disagreement}, 6.60\% of PubTabNet samples achieve TEDS$\geq$90\% while PA$<$80\%—cases where conventional evaluation would suggest success despite substantial misalignment. GAP reduces this disagreement rate to 5.75\%, a 12.9\% relative reduction, demonstrating that geometry-aware training improves not only raw accuracy but also consistency between structural and positional metrics.

Although TEDS improvements appear modest, this metric inherently masks pointer-level errors through partial credit scoring. Position Accuracy reveals the true improvement magnitude: 0.55\% absolute gain on PubTabNet corresponds to hundreds of corrected cell alignments. For applications like financial table extraction, where a single misaligned cell can invert values (Figure 5), such precision gains carry practical significance.

\subsection{Distance-Specific Error Analysis}

To quantify the effectiveness of GAP in addressing spatial locality patterns, we analyze pointer errors by Manhattan distance ranges before and after applying GAP.

\begin{figure}[t]
    \centering

    \begin{subfigure}[t]{0.35\linewidth}
        \centering
        \vspace{0pt}\includegraphics[width=\linewidth]{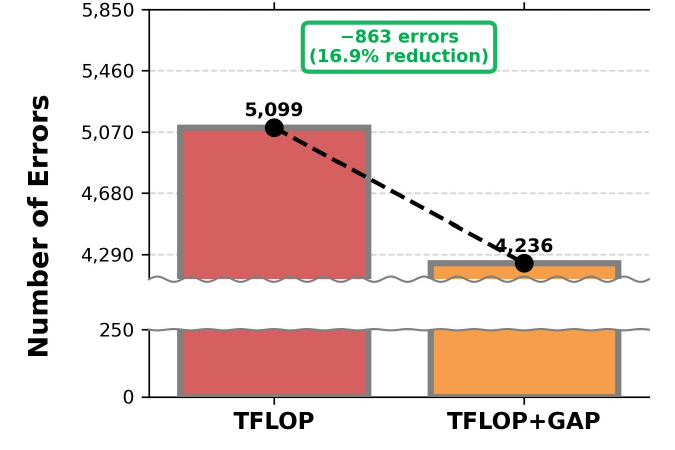}
        \caption{$d \in [1,2]$}
        \label{fig:sub1}
    \end{subfigure}
    \begin{subfigure}[t]{0.35\linewidth}
        \centering
        \vspace{0pt}\includegraphics[width=\linewidth]{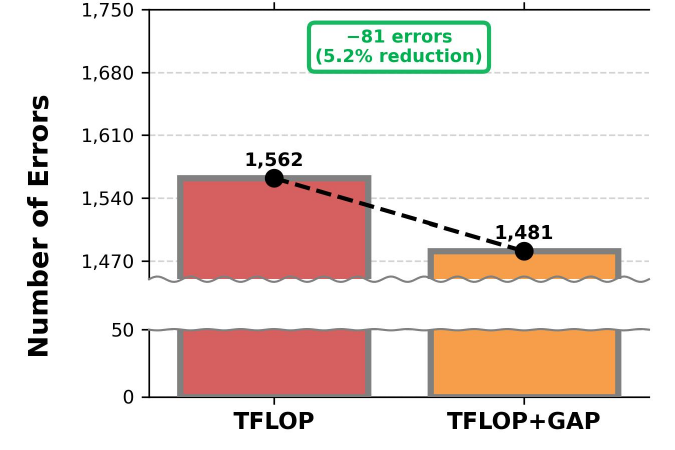}
        \caption{$d \in [3,4]$}
        \label{fig:sub2}
    \end{subfigure}

    \par\medskip 
    \begin{subfigure}[t]{0.35\linewidth}
        \centering
        \vspace{0pt}\includegraphics[width=\linewidth]{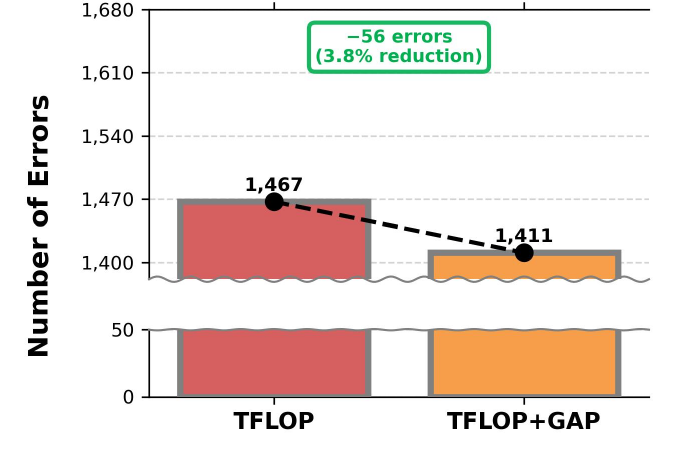}
        \caption{$d \in [5,8]$}
        \label{fig:sub3}
    \end{subfigure}
    \begin{subfigure}[t]{0.35\linewidth}
        \centering
        \vspace{0pt}\includegraphics[width=\linewidth]{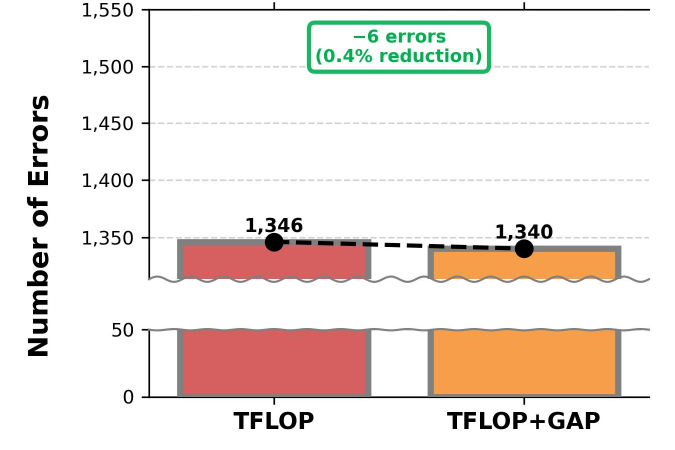}
        \caption{$d \in [9,20]$}
        \label{fig:sub4}
    \end{subfigure}

    \caption{Comparison of TFLOP and TFLOP+GAP performance across different $d$ values.}
    \label{fig:dist_errors}
\end{figure}

Figure~\ref{fig:dist_errors} shows that GAP reduces pointer errors across the evaluated local and mid-range distance bins, with the most pronounced gains occurring for adjacent-cell errors. Specifically, GAP reduces errors at $d \in [1,2]$ by 863 errors (16.9\% relative reduction), at $d \in [3,4]$ by 81 errors (5.2\% relative reduction), at $d \in [5,8]$ by 56 errors (3.8\% relative reduction), and at $d \in [9,20]$ by 6 errors (0.4\% relative reduction).

Most notably, the largest absolute and relative improvements occur at $d \in [1,2]$, where pointer networks typically struggle most with adjacent-cell confusions. The gains become smaller as the distance increases, indicating that GAP primarily targets local spatial confusions rather than uniformly affecting all error ranges. This pattern supports our hypothesis that GAP's distance-based reweighting concentrates learning signals where spatial confusions are most problematic. Due to space constraints, additional qualitative examples and failure case analysis are provided in the supplementary material.
  
\subsection{Table Complexity Analysis}

\begin{table}[t]
\centering
\caption{Performance by table complexity on PubTabNet test set}
\label{tab:complexity_analysis}
\footnotesize
\begin{tabular}{l|c|c||c|c}
\hline
\multirow{2}{*}{\textbf{Complexity}} & \multicolumn{2}{c||}{\textbf{TEDS}} & \multicolumn{2}{c}{\textbf{Position Acc}} \\
& \textbf{TFLOP} & \textbf{+GAP} & \textbf{TFLOP} & \textbf{+GAP}\\
\hline
Simple ($\leq$20) & 97.40 & \textbf{97.40} & 89.42 & \textbf{89.55} \\
Medium (21-50) & 97.06 & \textbf{97.09} & \textbf{92.15} & 92.14  \\
Complex (51-100) & 96.35 & \textbf{96.44} & 93.08 & \textbf{93.32}  \\
Very Complex ($>$100) & 95.12 & \textbf{95.25} & 91.08 & \textbf{92.26} \\
\hline
\end{tabular}
\end{table}

We analyze GAP performance across different table complexities to understand its robustness in various scenarios. Table~\ref{tab:complexity_analysis} reveals that GAP's benefits become more pronounced as table complexity increases. While TEDS improvements are relatively modest across all complexity levels (0.03--0.13\%), PA shows more substantial gains, particularly in very complex tables.

For simple tables ($\leq$ 20 cells), GAP achieves a 0.13\% improvement in PA (89.42\% → 89.55\%). Medium-complexity tables show minimal change (-0.01\%), while complex tables (51-100 cells) demonstrate a 0.24\% gain. Most notably, very complex tables ($>$ 100 cells) exhibit the largest PA enhancement with a 1.18\% increase (91.08\% → 92.26\%).

This robustness across varying complexities, combined with the lack of performance degradation in simpler cases, establishes GAP as a reliable enhancement whose benefits scale with recognition difficulty. Qualitative examples of these corrections are provided in the supplementary material.

\section{Conclusion}

In this work, we identify a fundamental mismatch in pointer-based TSR: while 79.6\% of errors occur between spatially adjacent cells, standard cross-entropy loss treats all negative candidates uniformly. This spatial blindness has been overlooked in existing approaches.

Our proposed Geometry-Aware Pointer (GAP) loss addresses this mismatch through distance-based reweighting that amplifies gradients for nearby cells. GAP requires no architectural modifications and adds zero inference cost. Extensive experiments on PubTabNet and SynthTabNet demonstrate state-of-the-art performance, with particularly substantial improvements in Position Accuracy across varying table complexities. Our findings suggest that careful analysis of error patterns can yield simple yet effective optimization strategies for structured prediction tasks.

\subsubsection*{Acknowledgmen}
This work was supported by (1) Information \& Communications Technology Planning \& Evaluation (IITP) grant funded by the Korea government (MSIT) (RS-2023-00259806), (2) Technology Development Program (TIPS) (RS-2024-00554500) funded by the Ministry of SMEs and Startups (MSS, Korea), and (3) Marine and Fisheries Deeptech R\&D Program funded by the Ministry of Oceans and Fisheries (MOF, Korea) through the Korea Institute of Marine Science \& Technology Promotion (KIMST) (RS-2026-25544055).

\bibliographystyle{splncs04}
\bibliography{main}

\end{document}